\documentclass{article}
\usepackage{spconf,amsmath,graphicx,hyperref}
\usepackage{times}  
\usepackage{helvet}  
\usepackage{courier}  

\title{Lotus: Efficient LLM Training by Randomized Low-Rank Gradient Projection with Adaptive Subspace Switching}
\name{Tianhao Miao\textsuperscript{1,*}, Zhongyuan Bao\textsuperscript{2,*}, Lejun Zhang\textsuperscript{3,*}\thanks{*Equal Contribution.}}
\address{
  \textsuperscript{1}Hong Kong Baptist University, School of Science, Hong Kong, China \\
  \textsuperscript{2}Fudan University, School of Data Science, Shanghai, China \\
  \textsuperscript{3}New York University, Tandon School of Engineering, New York, USA
}

\usepackage{algorithm}
\usepackage{algorithmic}

\usepackage{newfloat}
\usepackage{listings}

\usepackage{multirow}
\usepackage{booktabs}  
\usepackage{amsmath}
\usepackage{pifont}
\usepackage{latexsym}
\usepackage{colortbl}
\usepackage{amsmath,amssymb,amsthm}
\usepackage{bm}                  
\usepackage{xcolor}
\usepackage{makecell}

\begin{document}
\ninept
\maketitle
\begin{abstract}
Training efficiency in large-scale models is typically assessed through memory consumption, training time, and model performance. Current methods often exhibit trade-offs among these metrics, as optimizing one generally degrades at least one of the others. Addressing this trade-off remains a central challenge in algorithm design.  While GaLore enables memory-efficient training by updating gradients in a low-rank subspace, it incurs a comparable extra training time cost due to the Singular Value Decomposition(SVD) process on gradients. In this paper, we propose \textbf{Lotus}, a method that resolves this trade-off by simply modifying the projection process. We propose a criterion that quantifies the displacement of the unit gradient to enable efficient transitions between low-rank gradient subspaces. Experimental results indicate that Lotus is the most efficient method, achieving \textbf{a 30\% reduction in training time} and \textbf{a 40\% decrease in memory consumption} for gradient and optimizer states. Additionally, it outperforms the baseline method in both pre-training and fine-tuning tasks.
\end{abstract}


\begin{figure}[ht]
\begin{center}
\includegraphics[width=0.42\textwidth]{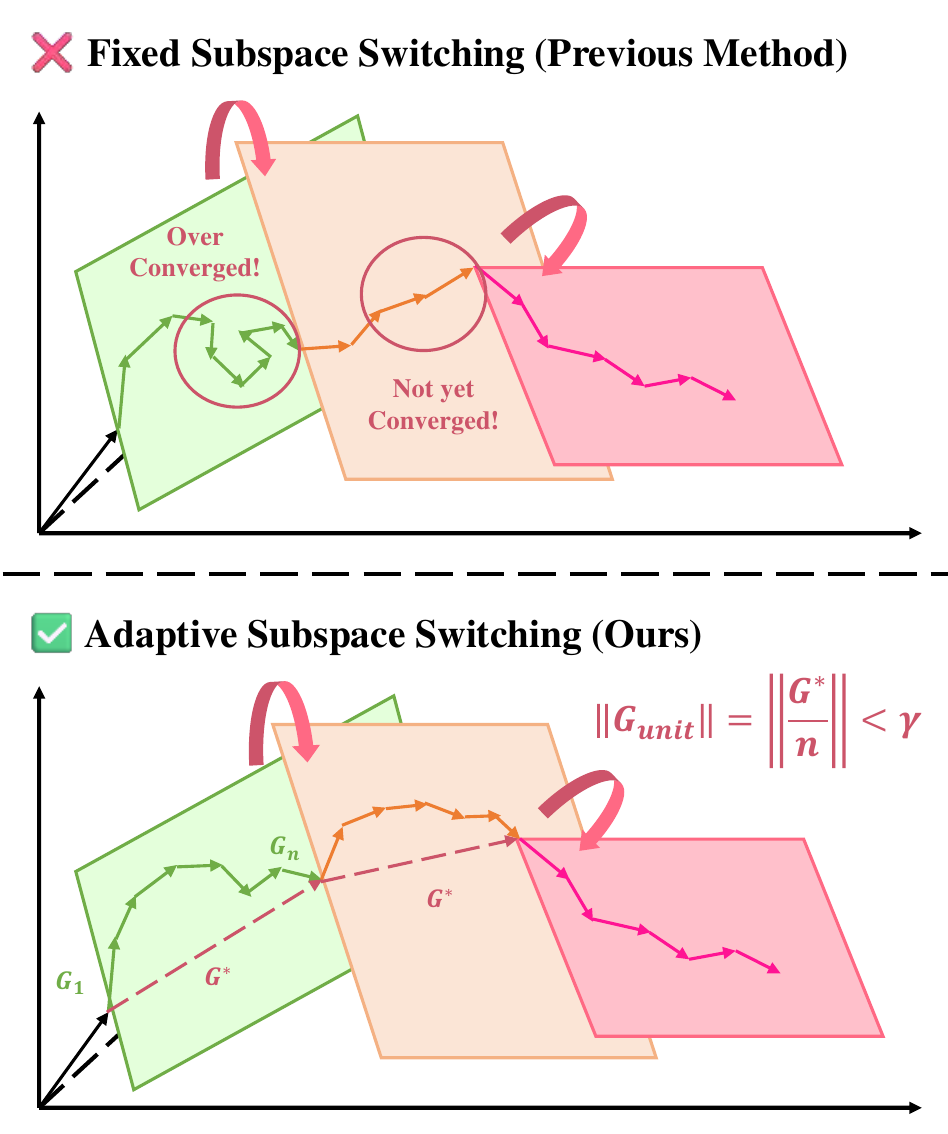}
  \caption{The comparison between previous method(e.g. GaLore) with fixed switching frequency and our greedy search strategy that updates the subspace adaptively. $G^*$ is the displacement of the unit gradient in a subspace. When the average displacement of unit gradient vector $G_{unit}$ is lower than $\gamma$, the subspace will be switched.}
  \label{fig:Algo-plot}
\end{center}
\end{figure}

\section{Introduction}
With the rapid advancement of large language models (LLMs), GPU memory requirements for training have substantially increased. Specifically, strategies such as using larger batch sizes and longer sequence lengths—which enhance model stability and generalization—have significantly elevated memory consumption. The memory footprint during LLM training typically comprises four components: model weights, gradients, optimizer states and activations. Researchers have proposed various memory-efficient techniques addressing different aspects mentioned above, including activation recomputation \cite{korthikanti2023reducing}, mixed-precision training, memory-efficient optimizers \cite{shazeer2018adafactor}, and parameter-efficient fine-tuning (PEFT). Among these methods, low-rank adaptations of model weights, such as Lora \cite{hu2022lora} and its variants \cite{lialin2023relora,zhang2023adalora}, have gained attention due to their reduced memory usage and comparable performance to full-parameter fine-tuning. Recent researches increasingly focus on leveraging low-rank matrix decomposition methods like SVD on model weights to improve performance \cite{meng2024pissa,lingam2024svft}.

From another perspective, the model weights may not inherently possess a low-rank structure, so as to the representation of learned new feature, leading to inferior model performance \cite{chen2025lora}. Compared with decomposing weight matrix, recent studies have proposed innovative approaches to activation \cite{chen2025lora}, gradient, and optimizer states with low-rank paradigm. \cite{hao2024flora} first finds that LoRA updates can be viewed as performing random projection on gradients. \cite{he2025gora} improves the random projection matrix and initializes weights based on gradient information. GaLore\cite{zhao2024galore} leverages low-rank gradients for subspace updates and optimizes weights by projection. Several GaLore variants \cite{huang2024galore,chen2024fira} progress on relieving memory further. 

Nevertheless, previous methodologies often exhibit inherent trade-offs between \textbf{memory, computation time and performance}, predominantly due to reliance on the frequent necessity to project gradients between full-rank and low-rank spaces. GaLore delivers marked memory efficiency gains but incurs roughly 2× the training time of LoRA due to frequent, costly SVD-based subspace updates. The need to specify both rank and update interval introduces tuning ambiguity and practical difficulty in deployment. Issues arise when gradients oscillate within a subspace due to saddle points or minima, while fixed intervals may prematurely or belatedly trigger subspace changes. Such weaknesses are visualized in Figure \ref{fig:Algo-plot}. In contrast, recent adaptive approaches, like shrinking rank strategies by \cite{refael2024adarankgrad}, incur complex calculations.

\textbf{Our Approach} To address the trade-offs and limitations mentioned, we conduct a theoretical analysis of gradient-matrix factorization during training. We find that most existing methods depend on SVD, whose computational time and memory requirements scale super linearly with matrix size. In addition, fixed-interval subspace switching may lead to premature gradient over-rotation, causing transitions before the current subspace is fully exploited. Preliminary experiments yield one key observation: although gradient magnitudes fluctuate during optimization, \textbf{unit-norm gradients tend to follow stable optimization trajectories}. Based on this, we hypothesize that the alignment between the unit-norm gradient and the subspace geometry is a primary factor influencing the effectiveness and efficiency of updates.

Thus, we propose \textbf{Lotus}, a faster and more memory-efficient algorithm. Lotus incorporates randomness into matrix factorization via randomized projections to approximate the decomposition, substantially reducing peak training memory. Building on a physics-inspired view of gradient descent, we design an adaptive update strategy governed by the Euclidean distance between low-rank unit gradients, switching subspaces dynamically when directional consistency indicates diminishing returns. The summary of our contribution is as follows:
\begin{enumerate}
    \item We introduce Lotus, a greedy search strategy to adaptively update subspace based on the average displacement of the unit gradient vector for better performance.
    \item Lotus can further save about 40\% memory consumption on gradient and optimizer states, and reduce 30\% time cost compared with GaLore. To performance, Lotus also exceeds related algorithms.
\end{enumerate}

\section{Related Works}

\noindent\textbf{Low-rank in Weight.}
LoRA shows the potential of memory-efficient learning for LLM with the intrinsic low-rank structure \cite{hu2022lora}. \cite{schotthofer2022low} constrains the rank of weight matrices to find "winning-ticket" dynamically in dense network. ReLoRA \cite{lialin2023relora} utilizes locally low-rank updates to train high-rank networks. RandLoRA \cite{albert2025randlora} updates a learned linear combinations of low-rank, non-trainable random matrices. Dora \cite{liu2024dora} decomposes weights to directional matrix and magnitude vector for fine-tuning. Further works improve the low-rank adapter with the help of SVD to the weights in the initialization of the adapters \cite{meng2024pissa}, only training the top-r singular values \cite{sun2024svfit}, or the corresponding coefficients \cite{lingam2024svft}. Lora-Null \cite{tang2025lora} finds the null space of representative activations and initializes the LoRA adapter with the pre‑trained weights in that null space. \cite{wang2024pmss} achieves high-rank updates with low costs by selecting skeletons from the pre-trained weight and learning a small matrix instead. \cite{jaiswal2024galore} unifies weight compression and memory-efficient fine-tuning as one to do adaptive low-rank weight projection.

\noindent\textbf{Low-rank in Optimizer States.}
Except for projecting the optimizer state to low-rank corresponding to the gradient structure, Fira \cite{chen2024fira} enables full-rank training under low-rank optimizer constraints by leveraging norm-based scaling and a gradient norm-growth limiter. Adapprox \cite{zhao2024adapprox} finds the key effect of top singular values in the second-order momentum of Adam and approximates it by random projection. Apollo \cite{zhu2024apollo} adopts low-rank scaling factors by projecting gradients randomly, and channel-wise updates instead of element-wise in Adam to reduce optimizer memory cost. Alice \cite{gong2025towards} leverages Fisher information matrix approximation to Adam for saving memory. COSMOS \cite{liu2025cosmos} merges full‑second‑order on dominant gradient subspace and Newton‑Schulz approximation on residual directions.

\noindent\textbf{Low-rank in Gradient.}
Flora down-projects gradients by random matrix and up-projects low-rank gradients for optimization, and GoRA improves Flora by adaptive random projection matrix and gradient-related initialization \cite{hao2024flora,he2025gora}. GaLore \cite{zhao2024galore} factorizes gradients based on SVD and optimizes full-rank weights after projecting the low-rank update back to the original space, and its variants try to use less memory and relieve the complexity by random projection or layer-wise adaptation to low-rank gradients \cite{huang2024galore}. Adarankgrad \cite{refael2024adarankgrad} uses adaptive subspace based on lower intrinsic rank with training process. \cite{torroba2025duality} leverages linear gradient transformations meeting Kronecker-factored as equivalence to a linear adapter. \cite{liang2024memory} uses online PCA to update the projection matrix in subspace learning. \cite{chang2025elalora} dynamically prunes and expands ranks based on gradient-derived importance scores. \cite{chen2025lora} approximates matrix multiplication with only critical rows and columns and improves back propagation. GaLore 2 and Galore+ accelerate the projection process, but no proof and time complexity analysis are provided \cite{su2025galore}.
\begin{table*}[htbp]
\centering
\caption{
We compare the performance of several low-rank training algorithms with Lotus by pre-training LLaMA models of varying sizes on the C4 dataset. Here, \textit{r} denotes the rank of the low-rank factorization, and \textit{$d_{\text{model}}$} denotes the hidden states dimension of each model size.}
\tabcolsep=0.045\linewidth
\begin{tabular*}{0.91\linewidth}{@{\hspace{0.85em}}lcccc}
\toprule
\textbf{Method} & \textbf{60M} & \textbf{130M} & \textbf{350M} & \textbf{1B} \\
\midrule
Full Rank     & 34.06(0.36G) & 25.08(0.76G) & 18.80(2.06G) & 15.56(7.80G) \\
\midrule
GaLore        & 34.88(0.24G) & 25.36(0.52G) & 18.95(1.22G) & 15.64(4.38G) \\
Low Rank      & 78.18(0.26G) & 45.51(0.54G) & 37.41(1.11G) & 142.53(3.62G) \\
LoRA          & 34.99(0.36G) & 33.92(0.80G) & 25.58(1.76G) & 19.21(6.17G) \\
ReLoRA        & 37.04(0.36G) & 29.37(0.80G) & 29.08(1.76G) & 18.33(6.17G) \\
AdaRankGrad   & \underline{34.24(0.21G)} & \underline{25.22(0.50G)} & \underline{18.91(1.11G)} & \textbf{14.71(3.62G)} \\
Lotus         & \textbf{33.75(0.23G)} & \textbf{24.87(0.51G)} & \textbf{18.91(1.19G)} & \underline{15.33(4.20G)} \\
\midrule
\textit{r} / \textit{$d_{\text{model}}$} & 128 / 256 & 256 / 768 & 256 / 1024 & 512 / 2048 \\
Training Tokens & 1.1B & 2.2B & 6.4B & 13.1B \\
\toprule
\end{tabular*}
\label{pretrain_results}
\end{table*}

\section{Methodology}

\subsection{Adaptive Subspace Switching}
Refreshing the orthogonal projector with the latest full-rank gradient realigns the low-rank basis with the its current dominant directions, so the top r singular vectors reclaim energy that had drifted outside the stale subspace; the Frobenius norm of the compressed gradient therefore “jumps back up.” Yet because different spectral components of the gradient drift at different speeds, a fixed update frequency both wastes compute on already stable directions and allows fast-moving ones to leak energy before the next refresh. An adaptive switching schedule throttles and accelerates the process according to subspace drift to solve this imbalance.

To quantify how much displacement such a schedule can preserve, consider the ideal scenario in which every projected gradient step points in exactly the same direction; then the cumulative displacement after $k$ steps is
\begin{equation}
  \label{eq:didea2}
  D_{\text {ideal }}=\left\|\sum_{i=0}^{k-1}-\alpha \hat{g}_{t-i}\right\|_2= \alpha\left\|\sum_{i=0}^{k-1} \hat{g}_{t-i}\right\|_2 
\end{equation}
In the “best‑aligned” case where all $\hat{g}$ are parallel and have unit norm, $D_{\text {ideal }} \approx k \alpha$, where $\alpha$ is the learning rate. Then the actual displacement would be:
{\fontsize{10pt}{13pt}\selectfont \begin{equation}
\begin{aligned}
 D_{\text{actual}} = \left\| \sum_{i=0}^{k-1} \Delta w_{t-i} \right\|_2= \left\| \sum_{i=0}^{k-1} -\alpha P_k \hat{g}_{t-i} \right\|_2
\label{eq:actual_displacement}
\end{aligned}
\end{equation}}
Then, we define the path‑efficiency ratio:
\begin{equation}
  \label{eq:trigger}
  \rho_t=\frac{D_{\text {actual }}}{D_{\text {ideal }}}=\frac{\left\|\sum_{i=0}^{k-1} P_k \hat{g}_{t-i}\right\|_2}{\left\|\sum_{i=0}^{k-1} \hat{g}_{t-i}\right\|_2} \in[0,1]  
\end{equation}
when $\rho_t \approx 1$, the gradients remain confined within a narrow directional cone, indicating that the current subspace$\ P_k $ is sufficiently representative for optimization. If $\rho_t \ll 1$, significant cancellation occurs between successive steps, suggesting that the gradients exhibit substantial directional variation or extend beyond the span of the subspace $\ P_k $. Lotus adaptively switches the subspace when $\rho_t<\gamma$ and $t-t_{\text {last }} \geq T_{\min }$, with threshold $\gamma \in(0,1)$. Noticing that we set a minimum interval condition, constraint $t-t_{\text {last }} \geq T_{\min }$ is imposed to prevent excessive subspace switches during the initial noisy phase of optimization. Especially, $k = 1$ means that $\rho_t$ reduces to the single-step displacement-gradient ratio.

\noindent\textbf{Lemma 3.1} (one‑step projected decrease) \textit{If $\rho_t \geq \rho$ and the loss has standard $L$-smoothness. We can apply the standard upper bound for L-smooth functions to the subspace projected update rule, then:}
{\fontsize{9pt}{13pt}\selectfont \begin{equation}
\begin{aligned}
  \label{eq:proposition1}
 \mathcal{L}\left(w_{t+1}\right) &\leq \mathcal{L}\left(w_t\right)-\alpha \rho^2\left\|g_t\right\|_2^2+\frac{1}{2} \alpha^2 L\left\|g_t\right\|_2^2  
\end{aligned}
\end{equation}}

Where $w_t \in \mathbb{R}^d$ is the parameter vector in iteration $t$ and $g_t=\nabla \mathcal{L}\left(w_t\right)$ is the gradient of the loss function at step $t$.

\begin{algorithm}
\caption{Lotus Algorithm}
\begin{algorithmic}
\STATE \textbf{Input:} Full-rank gradient $G_F \in \mathbb{R}^{m \times n}$; avg. unit gradient displacement threshold $\gamma$; verifying gap $\eta$ 
\STATE \textbf{Initialize:} Project count $T \gets 0$
\IF{Initialization or Subspace Switch}
    \STATE $O_G \gets \textsc{Efficient Low-Rank Project}(G_F)$
    \STATE $G_{\text{init}} \gets O_G \cdot G_F$
    \STATE $d_{\text{init}} \gets \textsc{Normalize}(G_{\text{init}})$
    \STATE $T \gets 1$
\ENDIF

\STATE $G_{\text{cur}} \gets O_G \cdot G_F$
\STATE $d_{\text{cur}} \gets \textsc{Normalize}(G_{\text{cur}})$
\STATE $T \gets T + 1$

\IF{$T \bmod \eta = 0$}
    \STATE $\Delta d \gets d_{\text{cur}} - d_{\text{init}}$
    \STATE $\lVert \bar{d} \rVert \gets \lVert \Delta d \rVert / T$ \hfill \textit{(avg. displacement)}
    \IF{$\lVert \bar{d} \rVert < \gamma$}
        \STATE Trigger Subspace Update
    \ENDIF
\ENDIF
\end{algorithmic}
\label{algo:lotus}
\end{algorithm}

\noindent\textbf{Theorem 3.2} (faster convergence with adaptive policy) \textit{Let $N_{\text {fix}}$ and $N_{\text {ada}}$ denote the number of iterations required by the fixed and adaptive step size policies, respectively, to achieve the gradient tolerance condition $\sum_{t=0}^{N-1} G_t \leq \epsilon$, where $G_t=\left\|g_t\right\|_2^2$ and the step size constraint $\alpha<2 \rho_{\text {fix }}^2 / L$. Then the following inequality holds:
\begin{equation}
  \label{eq:proposition0}
  N_{\mathrm{ada}} \leq \frac{c_{\mathrm{fix}}}{c_{\mathrm{ada}}} \frac{k}{T} N_{\mathrm{fix}}<N_{\mathrm{fix}} 
\end{equation}}

This result demonstrates that Lotus's adaptive subspace switching achieves the same convergence criterion with strictly fewer iterations compared to the fixed policy, highlighting its efficiency.

\begin{table*}[ht]
\centering
\caption{Evaluating Lotus on the GLUE benchmark. We compare Lotus with various memory-efficient training methods and report the average metrics. We set the threshold $\gamma$=0.01 and verifying gap $\eta$=50 as our baseline throughout the fine-tuning tasks.}
\tabcolsep=8.8pt
\begin{tabular*}{\linewidth}{@{\hspace{0.2em}}lcccccccccc}
\toprule
\textbf{Method} & \textbf{Memory} & \textbf{CoLA} & \textbf{STS-B} & \textbf{MRPC} & \textbf{RTE} & \textbf{SST2} & \textbf{MNLI} & \textbf{QNLI} & \textbf{QQP} & \textbf{Avg} \\
\midrule
Full Fine-Tuning     &  747M & 62.24 & 90.92 & 91.30 & 79.42 & 94.57 & 87.18 & 86.28 & 92.28 & 86.28 \\
\midrule
LoRA (rank=4)          & 257M & 61.38 & 90.57 & 91.07 & 78.70 & 92.89 & 86.82 & 92.18 & \textbf{91.29} & 85.61 \\
GaLore (rank=4)        & 253M & 60.35 & 90.73 & 92.25 & 79.42 & 94.04 & \underline{87.00} & 92.24 & 91.06 & 85.89 \\
Apollo (rank=4)        & 251M & 59.75 & 89.89 & 90.74 & 74.00 & 93.11 & 85.62 & 91.70 & 89.38 & 83.87 \\
AdaRankGrad (rank=4)   & 202M & \underline{61.40} & \textbf{90.97} & \underline{92.60} & \underline{81.23} & \textbf{94.80} & 86.60 & \underline{92.50} & 90.40 & \underline{86.31} \\
Lotus (rank=4)       & 251M & \textbf{64.67} & \underline{90.79} & \textbf{93.14} & \textbf{83.39} & \underline{94.72} & \textbf{87.47} & \textbf{93.00} & \underline{91.06} & \textbf{87.28} \\
\midrule
LoRA (rank=8)          & 264M & 61.83 & 90.80 & 91.90 & 79.06 & 93.46 & 86.94 & 92.25 & \textbf{91.22} & 85.93 \\
GaLore (rank=8)        & 257M & 60.06 & 90.82 & 92.01 & 79.78 & 94.38 & \underline{87.17} & 92.20 & 91.11 & 85.94 \\
Apollo (rank=8)        & 251M & 60.63 & 90.08 & 90.51 & 74.36 & 93.34 & 85.90 & 92.20 & 88.98 & 84.50 \\
AdaRankGrad (rank=8)   & 237M & \underline{62.00} & \underline{90.89} & \underline{93.20} & \underline{81.23} & \underline{94.80} & 86.50 & \underline{92.60} & 89.70 & \underline{86.36} \\
Lotus (rank=8)         & 254M & \textbf{63.44} & \textbf{91.06} & \textbf{93.35} & \textbf{81.58} & \textbf{94.95} & \textbf{87.32} & \textbf{93.11} & \underline{91.15} & \textbf{86.99} \\
\bottomrule
\end{tabular*}
\label{glue_results}
\end{table*}

\subsection{Lotus Algorithm}
In this section, we introduce Lotus, a training strategy designed to simultaneously accelerate computation and reduce memory usage. Lotus uses a power-iteration-based randomized SVD to markedly accelerate the gradient projection step. In addition, it incorporates a novel, more flexible path‑efficient switching policy: we define the path efficiency $\rho_t$ of the accumulated gradient displacement, and whenever $\rho_t$ drops below a preset threshold while the time since the previous switch exceeds $T_{\min }$, the algorithm triggers a subspace recomputation. The details are in Algorithm \ref{algo:lotus}. This mechanism guarantees that, compared to fixed-interval subspace switching, the adaptive strategy reaches the same gradient threshold in fewer iterations, thereby achieving faster convergence. The verifying gap $\eta$ should be set within 25-100 steps and the threshold $\gamma$ should be set within the 0.005-0.02 range to avoid too frequent or few updates.

\section{Experiments}
\begin{figure*}[h]
\begin{center}
\includegraphics[width=0.94\textwidth]{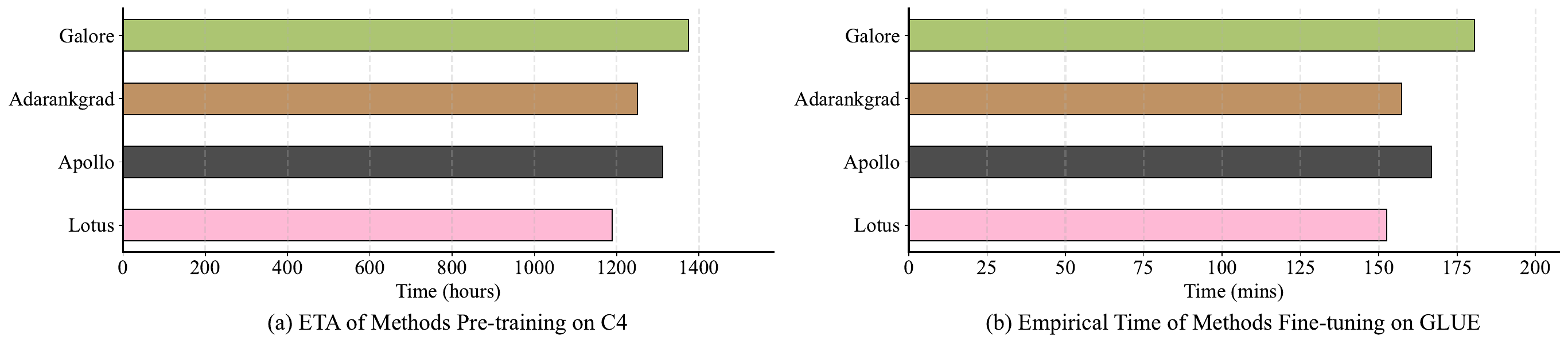}
  \caption{We visualize training time consumption comparsion between Galore, Adarankgrad, Apollo and Lotus on both pre-training and fine-tuning tasks. (a) shows the ETA of pre-training LLaMA-type 3B models in C4. (b) shows the average time cost on 8 GLUE tasks. Lotus is the most effective method in terms of computational time efficiency. (\textbf{Less is better})}
  \label{fig:eta}
\end{center}
\end{figure*}

\noindent\textbf{Implementation Details} We utilize GaLore \cite{zhao2024galore} as our codebase for model training and evaluation. All model architectures involved in the experiment are consistent with GaLore. The data format in training and validation is BF16. We tune the hyper-parameters needed in the experiments to achieve optimal performance.

\subsection{Pre-Training and Fine-tuning}
To evaluate the effectiveness of Lotus, we pre-train LLaMA models of varying sizes on the C4 dataset—a widely used cleaned version of the Common Crawl corpus, following GaLore’s experimental settings and using perplexity as the primary metric The corresponding experimental results are reported in Table \ref{pretrain_results}. All pre-training experiments here use NVIDIA H100 GPUs. 


We fine-tune RoBERTa-Base model on 8 GLUE tasks to compare the results with full rank fine-tuning, Lora, GaLore and AdaRankGrad, showing the results in Table \ref{glue_results}. We report the overall (matched and mismatched) accuracy for MNLI, Matthew’s correlation for CoLA, Pearson correlation for STS-B, F1 score for MRPC, and accuracy for other tasks. Lotus surpasses previous methods on most tasks, achieving higher average scores while reducing memory costs. All fine-tuning experiments use NVIDIA RTX 4090 GPUs.

\subsection{Training Time Efficiency and Performance Comparsion}
We compare the estimated time of arrivals (ETA) for pre-training LLaMA-type 3B models using an 8-bit optimizer with layer-wise weight updates on a single NVIDIA RTX 4090 GPU, following Galore and the average time cost on the fine-tuning tasks as shown in Figure \ref{fig:eta}. Our method demonstrates significant time savings compared to GaLore, AdaRankGrad, and Apollo. Additionally, Lotus shows faster subspace update frequency than GaLore in Table \ref{tab:time_saving}. These results indicate that Lotus is the simplest yet most effective method in terms of computational efficiency. We also quantify the contribution of randomized SVD(rSVD) and adaptive subspace switching (AdaSS) as shown in Table \ref{tab:ablation}. This shows that rSVD closely matches the exact-SVD variant at the same rank, and most of the gain comes from our adaptive subspace update. 

\newcommand{\gain}[1]{\scriptsize\raisebox{0.25ex}{$\uparrow$}{#1}}
\begin{table}[h]
\caption{The comparison of GaLore and Lotus on update frequency in GLUE benchmark.}
\centering
\tabcolsep=0.028\linewidth
\begin{tabular}{lccc}
\toprule
\textbf{Method}   & \textbf{\makecell{Subspace\\Account}} & \textbf{\makecell{Subspace Switching\\Frequency}}  \\ 
\midrule
GaLore (rank=4) & 3536  & 1.6 \\
Lotus (rank=4)  & \textbf{11614} & \textbf{6.5}~\gain{306\%}  \\ 
\midrule
GaLore (rank=8) & 3544  &  1.6    \\
Lotus (rank=8)  & \textbf{11736} & \textbf{6.3}~\gain{320\%} \\ 
\bottomrule
\end{tabular}
\label{tab:time_saving}
\end{table}

\begin{table}[htbp]
\caption{Component-wise evaluation of SVD, rSVD, and rSVD with AdaSS, illustrating how each component affects results.}
\centering
\tabcolsep=0.065\linewidth
\begin{tabular}{cccc}
\toprule
        \textbf{Rank}     & \textbf{rSVD} & \textbf{AdaSS} & \textbf{Avg}  \\ 
\midrule
4 &  &   & 85.89 \\
4  & \checkmark &   & 85.89 \\
4  & \checkmark & \checkmark &  \textbf{87.28} \\ 
\midrule
8 &  &   & 85.94 \\
8  & \checkmark &   & 86.07 \\
8 & \checkmark & \checkmark &  \textbf{86.99} \\ 
\bottomrule
\end{tabular}
\label{tab:ablation}
\end{table}

\section{Conclusion}
We introduce Lotus, an adaptive low-rank update algorithm that delivers a \textbf{30\%} training speedup and a \textbf{40\%} reduction in memory, while achieving better convergence than full-rank pretraining, fine-tuning and recent memory-efficient methods. Guided by the hypothesis that alignment between unit-gradient displacement and subspace geometry governs update efficiency, Lotus adopts a physics-inspired view of gradient descent: it tracks the Euclidean distance between low-rank unit gradients and dynamically switches subspaces when directional consistency degrades. We provide theoretical analysis and extensive experiments showing that Lotus effectively balances memory, computation time, and performance for LLM training, offering a practical tool for efficient large-model optimization.

\bibliographystyle{IEEEbib}
\bibliography{strings,main}

\end{document}